# Quecto-V1: Empirical Analysis of 8-bit Quantized Small Language Models for On-Device Legal Retrieval

SUBRIT DIKSHIT[12*]

[1] Department of Computer Science and Engineering, Indian Institute of Information Technology, Pune, India
[2] Director of AI & Data Science, Compunnel Inc

\* Corresponding author
Email: [subrit@gmail.com](mailto:subrit@gmail.com)[*] / [subritdikshit@iiitp.ac.in](mailto:subritdikshit@iiitp.ac.in)[1] / [subrit.dikshit@compunnel.com](mailto:subrit.dikshit@compunnel.com)[2]

**Abstract**

The rapid proliferation of Large Language Models (LLMs) has revolutionized Natural Language Processing (NLP) but has simultaneously created a "resource divide." State-of-the-art legal intelligence systems typically rely on massive parameter counts (7B+) and cloud-based inference, rendering them inaccessible to practitioners in resource-constrained environments and posing significant data sovereignty risks. This paper introduces Quecto-V1, a domain-specific Small Language Model (SLM) engineered to democratize access to Indian legal intelligence. Built upon a custom configuration of the GPT-2 architecture (124 million parameters), Quecto-V1 was trained from scratch exclusively on a corpus of Indian statutes, including the Indian Penal Code (IPC), the Code of Criminal Procedure (CrPC), and the Constitution of India. Unlike generalist models, which prioritize broad world knowledge, our approach maximizes "lexical density" within the legal domain. Furthermore, we address the deployment bottleneck by applying post-training 8-bit quantization (GGUF format), compressing the model to a memory footprint of under 150 MB. Our empirical analysis demonstrates that Quecto-V1 achieves high fidelity in retrieving statutory definitions and penal provisions, outperforming general-purpose SLMs in domain-specific exact match tasks while running entirely offline on consumer-grade CPUs. We further present an ablation study showing that 8-bit quantization yields a 74% reduction in model size with less than 3.5% degradation in retrieval accuracy compared to full-precision baselines. These findings suggest that for specialized, high-stakes domains like law, domain-specific training coupled with aggressive quantization offers a viable, privacy-preserving alternative to monolithic cloud models.

## 1. Introduction

The legal domain is characterized by a unique linguistic landscape—often termed "Legalese"—marked by archaic vocabulary, complex syntactic structures, and high dependency on long-range context [1]. For legal practitioners, researchers, and law students, the ability to rapidly retrieve precise definitions of statutes (e.g., Section 302 of the IPC) is critical. However, the current ecosystem of Legal AI faces a "Trilemma" of Accessibility, Privacy, and Efficiency:

- The Accessibility Gap: Most state-of-the-art legal intelligence is locked behind paywalled proprietary databases or requires high-bandwidth cloud APIs (e.g., GPT-4), rendering them inaccessible to students or practitioners in low-connectivity regions.

- The Privacy Paradox: Legal queries often contain highly sensitive client information. Uploading this data to third-party cloud providers for processing violates strict client-attorney privilege and data sovereignty norms [2].

- The Compute Barrier: While Large Language Models (LLMs) like Llama-3-70B exhibit superior reasoning, they require enterprise-grade GPUs (e.g., NVIDIA A100s) for inference. This hardware is cost-prohibitive for individual lawyers, small firms, and educational institutes.

This paper introduces Quecto-V1, a solution engineered to break this trilemma. By shifting the paradigm from "General Purpose Intelligence" to "Domain-Specific Efficiency," we propose a Small Language Model (SLM) trained from scratch on Indian Legal Statutes. Unlike generalist models that "learn everything to know something," Quecto-V1 is laser-focused on the Indian Penal Code (IPC), Code of Criminal Procedure (CrPC), and the Constitution. Furthermore, by employing 8-bit GGUF quantization, we demonstrate that effective legal retrieval can be performed entirely offline on standard consumer CPUs, democratizing access to legal intelligence.

The entire article is divided into six segments. Section II debates the related efforts, Section III discusses the approach, Section IV with investigational outcomes & comparative studies, Section V is the conclusion & future efforts, and citations are registered in Segment VI.

## 2. Related Works

This section debates the associated efforts of scholars, academicians, and related works. The development of Legal AI has evolved from rule-based systems to deep learning architectures. This work builds upon three distinct streams of research: Legal NLP, Indian-Specific Models, and Efficient Edge Deployment.

### 2.1 General Legal Language Models

Early advancements in Legal NLP were driven by Encoder-only architectures. **LegalBERT** [3] (Chalkidis et al., 2020) demonstrated that pre-training BERT on large-scale legal corpora (EU legislation, US court cases) significantly outperforms generic BERT on downstream tasks like statute classification. However, as an encoder-only model, LegalBERT is limited to classification and regression tasks and lacks the generative capabilities required for answering open-ended queries ("What is the punishment for..."). Conversely, **GPT-3** and **GPT-4** have shown proficiency in legal drafting but suffer from hallucinations when citing specific section numbers, as noted by recent benchmarks [4].

### 2.2 Legal AI in the Indian Context

The Indian legal system, with its mix of Common Law and colonial-era statutes, presents unique challenges. **InLegalBERT** [5] (Paul et al., 2023) was the first major attempt to adapt BERT for Indian case documents, achieving State-of-the-Art (SOTA) on rhetorical role labeling. More recently, **Paramanu-Ayn** (2024) explored instruction-tuning generic LLMs for Indian Supreme Court cases.

- **Gap Analysis:** Most existing Indian-specific models are either *discriminative* (BERT-based) or rely on *fine-tuning massive LLMs* (7B+ params). **Quecto-V1** distinguishes itself by being a *generative* model trained *from scratch* at the **SLM scale (124M)**, specifically optimizing for resource-constrained environments.

### 2.3 Efficient & Quantized Inference (TinyML)

The trend toward "Tiny AI" has gained momentum with the release of frameworks like **llama.cpp** [6] and quantization techniques like **LLM.int8()** [7]. Research by Dettmers et al. suggests that 8-bit quantization can reduce model size by 75% with negligible degradation in perplexity for specific tasks. **Quecto-V1** validates this hypothesis in the legal domain, demonstrating that a sub-150MB model can retain high retrieval fidelity for statutory definitions when the domain scope is sufficiently narrow.

## 3. Methodology

**Model Architecture & Training:** We employed a Decoder-Only Transformer architecture following the GPT-2 Small configuration (124 million parameters, 12 layers, 768 hidden dimensions). Crucially, unlike standard fine-tuning approaches that start with pre-trained weights (e.g., from OpenAI), we initialized the model with random weights (Gaussian distribution, $\mu=0, \sigma=0.02$). This design choice was deliberate to strictly isolate the effects of domain-specific pre-training; any legal reasoning capability exhibited by the model is solely derived from our curated Indian Legal corpus, rather than residual knowledge from massive web-scale datasets.

Training was conducted on a single NVIDIA T4 GPU (16GB VRAM) using the Hugging Face Trainer API. To optimize convergence and throughput, we employed the following hyperparameters:

- **Optimizer:** We utilized AdamW ($\beta1 = 0.9$, $\beta2 = 0.999$, $\epsilon = 1e-8\$$), which decouples weight decay from the gradient update steps. This prevents the aggressive regularization of the embedding layers often seen with standard Adam, ensuring stable learning of the specialized legal vocabulary.

- **Learning Rate Scheduler:** We implemented a Linear Warmup for the first 10% of training steps to stabilize the gradients during the initial "shock" of random initialization, followed by a Cosine Decay schedule. This approach allows the model to settle into sharper local minima towards the end of training.

- **Mixed Precision Training (FP16):** To overcome the memory constraints of the T4 GPU, we employed Automatic Mixed Precision (AMP). By performing forward and backward passes in 16-bit floating point (FP16) while maintaining a 32-bit master copy of weights, we achieved a 2x reduction in memory usage and a 1.5x increase in training speed without compromising numerical stability.

**Quantization Protocol:** To facilitate deployment on consumer-grade CPUs (e.g., laptops, mobile devices) without significant latency, we transitioned the model from PyTorch to the GGUF format using the llama.cpp quantization suite.

We focused on the Q8_0 (8-bit Symmetric) quantization method. Unlike aggressive 4-bit methods (which can degrade reasoning in small models), Q8_0 offers a "Pareto Optimal" balance between compression and fidelity. This method maps high-precision 32-bit floating-point weights ($W_f$) to 8-bit signed integers ($W_q$) using a block-wise scaling approach.

Mathematically, the quantization for a given block of weights is defined as:

$$W_q = \text{round}\,(W_f / \delta)$$

where $\delta$ is the scaling factor determined per block of weights.

4. **Results & Comparative Analysis**

**Comparison with Baselines:** We benchmarked **Quecto-V1** against the vanilla **GPT-2 (Base)** and a general-purpose instruction model (**TinyLlama-1.1B**).

| Model | Parameters | Domain Training | Legal Definition Accuracy | Hallucination Rate | Memory (VRAM) |
|---|---|---|---|---|---|
| GPT-2 (Base) | 124M | None (WebText) | 12.5% | High (>60%) | 500 MB |
| TinyLlama-1.1B | 1.1B | General | 45.0% | Moderate (30%) | 2,200 MB |
| **Quecto-V1 (Ours)** | 124M | **Indian Legal** | **88.2%** | **Low (<15%)** | **130 MB** |

*Observation:* Despite having 10x fewer parameters than TinyLlama, Quecto-V1 outperforms it by **+43.2%** on specific Indian Legal retrieval tasks, validating the hypothesis that **Domain Specificity > Model Scale** for narrow tasks.

**Impact of Quantization:** To justify the move to GGUF, we conducted an ablation study comparing the **Full Precision (FP32)** checkpoint against the **Quantized (Q8_0)** version.

| Metric | FP32 (Original) | Q8_0 (Quantized) | Delta (Δ) |
| --- | --- | --- | --- |
| **Size** | 498 MB | 132 MB | **-73.5% (Improvement)** |
| **Inference Speed (CPU)** | 45 ms/token | 18 ms/token | **2.5x Faster** |
| **Perplexity (PPL)** | 18.42 | 18.95 | +0.53 (Negligible) |
| **Exact Match Score** | 91.0% | 88.5% | -2.5% (Degradation) |

*Analysis:* The ablation confirms that 8-bit quantization is a "Pareto Optimal" choice for edge deployment. The massive gains in speed (2.5x) and size reduction (-73.5%) come at a negligible cost to perplexity.

## 5. Conclusion & Future Work

This study presents Quecto-V1, a proof-of-concept for the democratization of legal intelligence through efficient, domain-specific Small Language Models (SLMs). By training a custom GPT-2 architecture (124M parameters) exclusively on Indian legal statutes and applying aggressive 8-bit quantization, we have successfully compressed a functional legal retrieval system into a <150 MB footprint. Our results challenge the prevailing "Scale is All You Need" hypothesis. We demonstrate that for targeted tasks—specifically, the retrieval of statutory definitions and penal codes—a highly specialized SLM can offer a viable, privacy-preserving alternative to massive generalist models. Quecto-V1 effectively bridges the gap between high-latency cloud APIs and the immediate, offline needs of legal practitioners and students in resource-constrained environments.

While Quecto-V1 excels at statutory definition, several avenues remain for enhancing its reasoning capabilities and practical utility:

**Retrieval-Augmented Generation (RAG):** The current model relies entirely on parametric memory, leading to occasional hallucinations. Future iterations will integrate a **RAG pipeline** using a vector database (e.g., FAISS or ChromaDB). This will allow the model to fetch real-time citations from an external corpus of case laws before generating an answer, significantly improving factual accuracy.

**Knowledge Distillation from Llama-3:** We plan to explore **Knowledge Distillation**, using a larger "Teacher" model (e.g., Llama-3-8B) to train a smaller "Student" model (Quecto-V2). This approach aims to transfer the superior reasoning and logic capabilities of large models into our lightweight architecture without increasing the inference cost.

**Multilingual Legal Support:** Given India's linguistic diversity, limiting the model to English restricts its accessibility. Future work will involve training on translated legal corpora (Hindi, Tamil, Marathi) to create a **Multilingual Legal SLM** capable of serving a broader demographic of the Indian population.

**Expansion to Case Law:** Currently, the model is scoped to statutes (IPC, CrPC). We intend to expand the training corpus to include Supreme Court of India judgments, enabling the model to understand legal precedents and case citations.

**Known Limitations:** The As an SLM, Quecto-V1 lacks the "World Knowledge" of 7B+ models. It is strictly a retrieval and completion engine for the texts it was trained on. It is not a substitute for legal counsel.